\documentclass[10pt,twocolumn,letterpaper]{article}

\usepackage[pagenumbers]{cvpr} 

\usepackage[dvipsnames]{xcolor}
\usepackage{mathtools}

\definecolor{cvprblue}{rgb}{0.21,0.49,0.74}
\usepackage[pagebackref,breaklinks,colorlinks,citecolor=cvprblue]{hyperref}

\def\paperID{*****} 
\def\confName{CVPR}
\def\confYear{2024}

\title{Enhanced Structured State Space Models via Grouped FIR Filtering and Attention Sink Mechanisms}

\author{
Tian Meng $^1$$^\dagger$\hspace{2em} 
Yang Tao $^2$$^\dagger$\hspace{2em}
Wuliang Yin $^1$$^*$
\\
$^1$University of Manchester\hspace{2em}
$^2$Mettler Toledo Safeline 
\\
$^\dagger$Joint First Author\hspace{5em}
$^*$Project Lead
}

\begin{document}
\maketitle
\begin{abstract}
Structured State Space Models (SSMs) have emerged as compelling alternatives to Transformer architectures, offering linear-time complexity and superior performance in various sequence modeling tasks. Despite their advantages, SSMs like the original Mamba-2 face training difficulties due to the sensitivities introduced by the extended series of recurrent matrix multiplications. In this paper, we propose an advanced architecture that mitigates these challenges by decomposing A-multiplications into multiple groups and optimizing positional encoding through Grouped Finite Impulse Response (FIR) filtering. This new structure, denoted as Grouped FIR-enhanced SSM (GFSSM), employs semiseparable matrices for efficient computation. Furthermore, inspired by the "attention sink" phenomenon identified in streaming language models, we incorporate a similar mechanism to enhance the stability and performance of our model over extended sequences. Our approach further bridges the gap between SSMs and Transformer architectures, offering a viable path forward for scalable and high-performing sequence modeling.
\end{abstract}    
\section{Introduction}
\label{sec:intro}

In recent years, the field of deep learning has been revolutionized by the introduction of the Transformer architecture and its attention mechanism \cite{vaswani_attention_2017,devlin_bert_2019,dosovitskiy_image_2020,dao_flashattention_2022,touvron_llama_2023}, which have become the backbone of most state-of-the-art models across various domains, including natural language processing (NLP), computer vision, and speech recognition. However, these architectures are not without limitations. Specifically, the quadratic time and memory complexity of the self-attention mechanism pose significant challenges when handling long sequences. To address these issues, several subquadratic-time architectures have been proposed, among which Structured State Space Models (SSMs) \cite{gu_efficiently_2022,sun_retentive_2023,gu_mamba_2024,dao_transformers_2024} have garnered considerable interest due to their linear-time complexity and promising performance. Mamba \cite{gu_mamba_2024} and its successor Mamba-2 \cite{dao_transformers_2024} represent significant strides in this direction, implementing SSMs that utilize selective state updates to efficiently process long sequences. Despite these advancements, Mamba-2 still encounters certain obstacles, primarily due to the sensitivity of the recurrent weight multiplications, which can complicate training. These sensitivities arise from a long sequence of multiplications, making the model potentially unstable and difficult to optimize effectively.

To alleviate these issues, we propose a novel architecture that reimagines the positional encoding mechanism within the SSM framework. Our innovation revolves around decomposing the recurrent multiplications into smaller, more manageable groups and employing Finite Impulse Response (FIR) filters \cite{saramaki_finite_1993,trimale_review_2017} to enhance positional encoding. This modification not only improves the sensitivity issues but also leveraging semiseparable matrices, which can be calculated with reduced complexity. Additionally, inspired by recent insights from the domain of streaming language models, we integrate an attention sinks mechanism \cite{xiao_efficient_2024,yu_unveiling_2024,cancedda_spectral_2024} within our architecture. Previous studies have highlighted the importance of retaining initial tokens as "sinks" that stabilize the attention mechanism over extended sequences. By incorporating a similar mechanism, our model benefits from improved stability and enhanced performance, particularly for tasks involving long-range dependencies.

We denote our new architecture as Grouped FIR-enhanced Structured State Space Model (GFSSM). GFSSM represents a dual-pronged improvement over traditional SSMs: first, by mitigating the training difficulties associated with long recurrent multiplications, and second, by incorporating attention sinks to handle extended sequences more effectively. In summary, our contributions can be outlined as follows:

\begin{enumerate}
    \item We introduce the Grouped FIR filter to optimize the positional encoding within SSMs, thereby addressing the sensitivity and training difficulties associated with long sequences of multiplications.
    \item We incorporate an attention sink mechanism to enhance model stability and performance over extended sequences, inspired by insights from streaming language models.
\end{enumerate}

This paper is structured as follows: Section 2 provides a detailed overview of related work, including foundational models and their limitations. Section 3 presents the mathematical formulation and architectural details of GFSSM. Section 4 outlines our experimental setup and results, followed by an in-depth analysis. Finally, Section 5 concludes the paper and discusses potential avenues for future work.

\section{Related Work}
\label{sec:related-work}

\subsection{Structured State Space Models (SSMs)}
Structured State Space Models (SSMs) \cite{gu_efficiently_2022,smith_simplified_2023,sun_retentive_2023,gu_mamba_2024,dao_transformers_2024} have attracted attention for their ability to offer linear-time complexity while maintaining competitive performance. Among these, the Mamba architecture \cite{gu_mamba_2024} represents a significant advancement. Mamba leverages selective state updates to efficiently handle long sequences, making it a viable alternative to Transformers. However, despite its linear complexity, Mamba's performance has been limited by the sensitivity and difficulty in training due to extended sequences of recurrent multiplications. Mamba-2 \cite{dao_transformers_2024} improved upon its predecessor by introducing a more structured approach, leveraging the State Space Duality (SSD) framework. By restricting the SSM parameters further and employing structured decompositions of semiseparable matrices, Mamba-2 achieved enhanced computational efficiency and performance. Nevertheless, the core challenge of training sensitivity due to recurrent dynamics remained an open problem.

\subsection{Finite Impulse Response (FIR) Filters}
FIR filters \cite{saramaki_finite_1993,trimale_review_2017} have been widely used in signal processing for their stability and straightforward implementation. Recent works have explored their potential in the context of machine learning, particularly for optimizing convolutional neural networks (CNNs) [6]. These filters can be tuned to emphasize specific frequencies, making them useful for various signal processing tasks. In the context of SSMs, deploying FIR filters can help mitigate sensitivity issues by effectively smoothing positional encodings.

\subsection{Attention Sinks}
The concept of attention sinks, as explored by \cite{xiao_efficient_2024,yu_unveiling_2024,cancedda_spectral_2024}, addresses some unique challenges in deploying large language models (LLMs) for streaming applications. By retaining key-value (KV) states of initial tokens, attention sinks stabilize the attention mechanism, thereby enabling LLMs to generalize to sequences much longer than those encountered during training. This insight is particularly relevant for our work, as it offers a potential solution to improve stability and performance over extended sequences in SSMs.
\section{Methodology}
\label{sec:methodology}
This section outlines the theoretical foundations and architectural innovations of the Grouped FIR-enhanced Structured State Space Model (GFSSM). We begin by revisiting the fundamental principles and equations that underpin our approach, followed by detailed explanations of the grouped FIR filtering and the integration of attention sinks. Finally, we discuss the integration of these components into the overall GFSSM architecture.

\subsection{Fundamentals of State Space Duality (SSD)}
In the Mamba-2 SSD framework, the forward pass is governed by the following equations:

\begin{equation}
    \begin{aligned}
        h_{t} &= A_{t} h_{t-1} + B_{t} x_{t} \\
        y_{t} &= C_{t}^{\top} h_{t},
    \end{aligned}
\end{equation}
where \( h_{t} \) denotes the hidden state at time \( t \), \( x_{t} \) is the input, and \( y_{t} \) represents the output. \( A_{t} \) is a scalar times identity matrix, denoted as \(a_t\), which simplifies the recurrent computation. \( B_{t} \) and \( C_{t} \) are matrices that define the state space dynamics, which can vary over time. By definition, \( h_0 = B_0 x_0 \). By induction, the \( h_{t} \) can be obtained as:

\begin{equation}
    \begin{aligned}
        h_t &= a_t a_{t-1} \cdots a_1 B_0 x_0 + a_t a_{t-1} \cdots a_2 B_1 x_1 + \cdots \\
        &\quad + a_t a_{t-1} B_{t-2} x_{t-2} + a_t B_{t-1} x_{t-1} + B_t x_t \\
        &= \sum_{s=0}^t a_{t:s}^\times B_s x_s,
    \end{aligned}
\end{equation}
where \( a_{t:s}^\times \) denotes the product of \( a \) terms from \( s+1 \) to \( t \). When multiplied by \( C_t \) to produce \( y_t \) and vectorized over \( t \in [\mathtt{T}] \), the matrix transformation form of SSD becomes:

\begin{equation}
    \begin{aligned}
        y_t &= \sum_{s=0}^t a_{t:s}^\times C_t^{\top} B_s x_s, \\
        y &= (L \circ C^{\top}B) x = Mx,
    \end{aligned}
\end{equation}
where \( L \) is represented as \( L_{ji} = a_j \cdots a_{i+1} \):

\begin{equation}
\label{eq:1ss}
    L =
    \begin{bmatrix}
        1 & & & & \\
        a_1 & 1 & & & \\
        a_2 a_1 & a_2 & 1 & & \\
        \vdots & \vdots & \ddots & \ddots & \\
        a_{T-1} \cdots a_1 & a_{T-1} \cdots a_2 & \cdots & a_{T-1} & 1 \\
    \end{bmatrix}.
\end{equation}

The potential limitation of the extended series of \( a_i \) multiplications arises when dealing with long text sequences. The matrix \( L_{ji} \) can decay to 0 or explode, resulting in numerical instability. Furthermore, any intermediate poorly conditioned \( a_i \) will have a prolonged adverse impact, propagating across subsequent tokens. For instance, if \( a_i \) is 0 at an initial position, all following states will remain 0. This extended multiplication series also complicates training, as noted in Mamba-2: higher precision for the main model parameters may be necessary due to the sensitivity of SSDs to recurrent dynamics. Parameter storage in fp32 can improve stability by mitigating these recurrent sensitivities.

\subsection{Grouped FIR Filtering}
In GFSSM, we address the sensitivity and training difficulties inherent in SSMs by optimizing the computation of the hidden states and refining the positional encoding mechanism. The sensitivity in training SSMs arises primarily from the extended series of recurrent multiplications. To mitigate this issue, we decompose recurrent multiplications into smaller, more manageable groups and employ Finite Impulse Response (FIR) filters to enhance the positional encoding. 

For simplicity, we illustrate the method using four groups with a 4th-order FIR filter. However, this can be easily generalized to any \( Q \) groups with an \( n \)-th order FIR filter. The new recurrent form is:

\begin{equation}
    \begin{aligned}
        s_{t} &= k_{0} B_{t} x_{t} + k_{1} B_{t-1} x_{t-1} \\
        &+ k_{2} B_{t-2} x_{t-2} + k_{3} B_{t-3} x_{t-3} \\
    \end{aligned}
\end{equation}

\begin{equation}
    \left\{
    \begin{split}
        h_{t}^{i} &= A_{t} h_{t-1}^{i} + s_{t}, \quad i = \text{mod}(t, Q) \\
        h_{t}^{i} &= h_{t-1}^{i}, \quad i \neq \text{mod}(t, Q)
    \end{split}
    \right.
\end{equation}

\begin{equation}
    y_t = C_t \sum_{i=0}^{Q-1} h_{t}^{i}
\end{equation}
where \( s_{t} \) is the filtered output, and \( k_{0}, k_{1}, k_{2}, k_{3} \) are learnable filter coefficients.

\begin{figure}[h]
  \centering
  \includegraphics[angle=-90,width=0.28\linewidth]{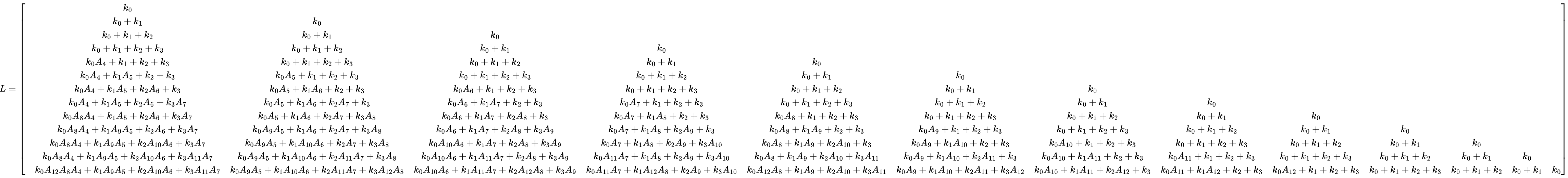}
  \caption{The matrix \( L \) for GFSSM}
  \label{fig:GFSSM_L}
\end{figure}

The core idea of GFSSM is to split the long input sequence into \( Q \) groups, each with its corresponding hidden state. This way, the recurrent multiplications of \( a_i \) are reduced to \( 1/Q \) of their original length, thereby mitigating their potential to cause numerical instability. Finally, the output \( y_t \) is computed by summing the contributions from all grouped states. One potential problem arises from splitting tokens into different groups: if tokens \( x_i \) and \( x_j \) belong to different groups, they will never be directly correlated. To address this issue, we introduce a computationally efficient solution by adding a FIR filter that processes the \( B_{t} x_{t} \) sequence. This enables tokens in different groups to be indirectly correlated, thereby preserving the overall coherence of the encoded sequence. To provide further clarity, we present the structure of matrix \( L \) for GFSSM, focusing on the first 13 elements, as shown in Figure \ref{fig:GFSSM_L}.
The application of FIR filters facilitates smoother transitions and correlations between grouped states, ultimately rendering the model more robust and easier to train. By combining grouped FIR filtering and robust matrix computations, GFSSM offers a viable solution for the inherent challenges of traditional SSM architectures, providing improved performance and stability over extended sequences.

\subsection{Integration of Attention Sink Mechanisms}
Inspired by the attention sink phenomenon observed in streaming language models, we incorporate a similar mechanism within GFSSM to improve stability and performance over extended sequences. The attention sink mechanism ensures that the initial tokens in a sequence serve as stable reference points, effectively acting as "sinks" that help maintain the model's focus and coherence over long dependencies. Additionally, it addresses the absence of initial states, thereby ensuring no information is lost from the initial \(a_0\) to \(a_Q\).

In our GFSSM architecture, we initialize the state using learnable prompts that serve as stand-ins for the first few tokens. This initialization is critical for ensuring meaningful state updates from the beginning of the sequence. Formally, we define the state updates incorporating the attention sink mechanism as follows, beginning with the initialization using learnable prompts. For initialization, we start with:
\begin{equation}
    \begin{split}
        h_{\text{init}}^{i} &= 0, \\
        y &= (L \circ C^{\top}B) 
        \begin{bmatrix}
            x_{-Q} \\
            x_{-Q+1} \\
            \vdots \\
            x_{-1} \\
            x_{0} \\
            x_{1} \\
            \vdots
        \end{bmatrix}
        + P \cdot \mathbf{0}
    \end{split}
\end{equation}
where the initial sequence \( \{ x_{-Q}, x_{-Q+1}, \ldots, x_{-1} \} \) consists of \( Q \) learnable prompts. For continued state updates, the state initialization incorporates the cached prior hidden state:
\begin{equation}
    \begin{split}
        h_{\text{init}}^{i} &= h_{-Q}^{i}, \\
        Y &= (L \circ C^{\top}B) 
        \begin{bmatrix}
            x_{-Q+1} \\
            x_{-Q+2} \\
            \vdots \\
            x_{0} \\
            x_{1} \\
            \vdots
        \end{bmatrix}
        + P 
        \begin{bmatrix}
            h_{-Q}^{1} \\
            h_{-Q}^{2} \\
            h_{-Q}^{3} \\
            h_{-Q}^{4}
        \end{bmatrix}
    \end{split}
\end{equation}

\begin{equation}
\resizebox{.8\hsize}{!}{$
    P = 
    \begin{bmatrix}
        A_{0} & 1      & 1      & 1 \\
        A_{0} & A_{1}  & 1      & 1 \\
        A_{0} & A_{1}  & A_{2}  & 1 \\
        A_{0} & A_{1}  & A_{2}  & A_{3} \\
        A_{4}A_{0} & A_{1}  & A_{2}  & A_{3} \\
        A_{4}A_{0} & A_{5}A_{1} & A_{2}  & A_{3} \\
        A_{4}A_{0} & A_{5}A_{1} & A_{6}A_{2} & A_{3} \\
        A_{4}A_{0} & A_{5}A_{1} & A_{6}A_{2} & A_{7}A_{3} \\
        A_{8}A_{4}A_{0} & A_{5}A_{1} & A_{6}A_{2} & A_{7}A_{3} \\
        A_{8}A_{4}A_{0} & A_{9}A_{5}A_{1} & A_{6}A_{2} & A_{7}A_{3} \\
        A_{8}A_{4}A_{0} & A_{9}A_{5}A_{1} & A_{10}A_{6}A_{2} & A_{7}A_{3} \\
        A_{8}A_{4}A_{0} & A_{9}A_{5}A_{1} & A_{10}A_{6}A_{2} & A_{11}A_{7}A_{3} \\
        A_{12}A_{8}A_{4}A_{0} & A_{9}A_{5}A_{1} & A_{10}A_{6}A_{2} & A_{11}A_{7}A_{3} \\
    \end{bmatrix},
$}
\end{equation}
where the initial sequence \( \{ x_{-Q+1}, x_{-Q+2}, \ldots, x_{0} \} \) is the preceding \( Q-1 \) input tokens, and \( h^{i}_{-Q} \) is the cached hidden state from the previous sequence.

This attention sink mechanism ensures that the initial tokens persist as stable anchors that maintain focus over the sequence's length, thus improving the robustness of our model. Combining grouped FIR filtering and the attention sink mechanism, our final GFSSM model effectively leverages semiseparable matrices within the broader SSD architecture, offering enhanced performance and stability over extended sequences. 

\subsection{Semiseparable Matrix Decomposition}
A crucial component of our new structure is the decomposition of the recurrent update matrix into multiple semiseparable matrices. This decomposition enables efficient computation while maintaining the linear complexity of state updates. We denote the decomposed matrix as \( L \), formulated as:

\begin{equation}
    L = k_{0} L_{0} + k_{1} L_{1} + k_{2} L_{2} + k_{3} L_{3},
\end{equation}
where each \( L_{i} \) represents a semiseparable matrix that captures the contributions from different stages of the FIR filter. Semiseparable matrices are characterized by a structure that allows fast multiplications, like Mamba-2, significantly reducing computational overhead.

\section{Conclusion and future work}
\label{sec:conclusion}
In this paper, we introduced the Grouped FIR-enhanced Structured State Space Model (GFSSM) as an innovative extension to traditional SSM architectures. Our approach addresses critical training challenges posed by extended series of recurrent matrix multiplications, which often lead to instabilities and difficulties in optimization. By decomposing these multiplications into more manageable groups and incorporating Finite Impulse Response (FIR) filters, we could potentially improve the robustness and trainability of the system. Additionally, inspired by the "attention sink" phenomenon observed in streaming language models, we integrated a similar mechanism into GFSSM to enhance stability and performance over extended sequences.

While we have outlined the theoretical foundations and architectural details of GFSSM, empirical validations are essential to demonstrate the model's practical advantages. Therefore, future work will entail comprehensive experiments to evaluate the performance of GFSSM across various sequence modeling tasks, including but not limited to natural language processing, speech recognition, and genomics.
{
    \small
    \bibliographystyle{ieeenat_fullname}
    \bibliography{references}
}


\end{document}